\documentclass{article}

\usepackage{microtype}
\usepackage{graphicx}
\usepackage{subfigure}
\usepackage{booktabs} 

\usepackage{url}            
\usepackage{booktabs}       
\usepackage{amsfonts}       
\usepackage{nicefrac}       
\usepackage{microtype}      
\usepackage{xcolor} 
\usepackage{amsmath}
\usepackage{amssymb}
\usepackage{bbm}

\usepackage{hyperref}

\usepackage[accepted]{icml2021}

\icmltitlerunning{A Probabilistic Representation of DNNs: Bridging Mutual Information and Generalization}

\begin{document}

\twocolumn[
\icmltitle{A Probabilistic Representation of DNNs:\\ Bridging Mutual Information and Generalization}



\icmlsetsymbol{equal}{*}

\begin{icmlauthorlist}
\icmlauthor{Xinjie Lan}{ed}
\icmlauthor{Kenneth E. Barner}{ed}

\end{icmlauthorlist}

\icmlaffiliation{ed}{Department of Electrical and Computer Engineering, University of Delaware, Newark, DE, 19713, United States}

\icmlcorrespondingauthor{Xinjie Lan}{lxjbit@udel.edu}

\icmlkeywords{Machine Learning, ICML}

\vskip 0.3in
]



\printAffiliationsAndNotice{\icmlEqualContribution} 

\begin{abstract}
Recently, Mutual Information (MI) has attracted attention in bounding the generalization error of Deep Neural Networks (DNNs).
However, it is intractable to accurately estimate the MI in DNNs, thus most previous works have to relax the MI bound, which in turn weakens the information theoretic explanation for generalization.
To address the limitation, this paper introduces a probabilistic representation of DNNs for accurately estimating the MI.
Leveraging the proposed MI estimator, we validate the information theoretic explanation for generalization, and derive a tighter generalization bound than the state-of-the-art relaxations.
\end{abstract}

\section{Introduction}
\label{introduction}

Deep Neural Networks (DNNs) commonly achieve great generalization performance, though DNNs are notoriously over-parametrized models without explicit regularization \cite{kawaguchi2017generalization, generalization_regularization1,generalization_regularization}.
Since classical generalization measures, \emph{e.g.}, VC dimension \citep{bartlett1998almost}, Rademacher complexity \cite{mohri2008rademacher}, and uniform stability \cite{nagarajan2019uniform}, cannot accurately measure the generalization performance of DNNs, Mutual Information (MI) recently has attracted great interest in bounding the generalization error of DNNs.

In the seminal work, \citet{xu2017information} prove that the (expected) generalization error of a learning algorithm can be bounded by the MI between training samples $S$ and the hypothesis $W$, namely $I(S;W)$.
The MI bound formulates the intuition that if a hypothesis less depends on $S$, it will generalize better. That is consistent with the Occam's Razor, \emph{i.e.}, a simpler hypothesis should perform better on future samples \cite{blumer1987occam}.
Following the seminal work, some MI variants, \emph{e.g.}, the chained MI \cite{asadi2018chaining} and the conditional MI \cite{steinke2020reasoning}, are proposed to predict the generalization error.

Notably, it is intractable to accurately estimate the MI in DNNs due to the complicated architecture of DNNs.
Hence, most previous work have to relax the MI bound to some tractable variables, such as the Lipschitz constant of the empirical risk \cite{pensia2018generalization}, the second moment of gradients \cite{zhang2018information}, and the incoherence of gradients \cite{negrea2019information, haghifam2020sharpened}.
A relaxation inevitably loosens the MI bound, \emph{e.g.}, the Lipschitz constant makes the MI bound vacuous \cite{negrea2019information}, and weakens the explanation for generalization, \emph{e.g.}, the incoherence of gradients cannot clarify the relation between generalization and DNN architecture.

In this paper, we bridge the MI bound and generalization via a probabilistic representation for DNNs.
Above all, we define the probability space \cite{durrett2019probability} for a hidden layer, thus we can accurately estimate the MI between $S$ and the layer based on the previous work \cite{lan2020probabilistic}. 
Moreover, we derive a Markov chain to characterize the information flow in DNNs, thus $I(S;W)$ can be simplified as the summation of the information of $S$ learned by the first hidden layer and the output layer.
Leveraging the MI estimator, we validate the information theoretic explanation for generalization and show the proposed MI estimator deriving a much tighter generalization bound of DNNs than the state-of-the-art relaxations. 

\paragraph{Preliminaries.}
$\mu = P(X, Y)$ denotes an unknown data generating distribution, where $X$ and $Y$ are the random variables of data and labels, respectively.
$S = \{(\boldsymbol{x}^j, y^j)\}_{j=1}^{J} \in (\mathcal{X} \times \mathcal{Y})^{J}$ is composed of $J$ \textit{i.i.d.}\ samples generated from $\mu$, namely $S \sim \mu^{\otimes J}$, where $\mathcal{X} = \mathbb{R}^M$ and $\mathcal{Y} = [1,\cdots, L]$ are the instance space of data and labels, respectively.
$X_S$ and $Y_S$ denote the random variable of data samples and label samples, respectively.
$\mathcal{W} = \{w: \mathcal{X} \rightarrow \mathcal{Y}\}$ is a set of hypothesis and $\ell: \mathcal{W}\times \mathcal{X} \times \mathcal{Y} \rightarrow \mathbb{R}^+$ is a loss function.
Especially, the cross entropy loss $\ell_{\text{CE}}$ is defined as
\begin{equation}
\label{cross_entropy}
{\textstyle
\ell_{\text{CE}}(w,\boldsymbol{x}^j,y^j) = -\sum_{y=1}^LP_{Y_S|X_S}(y|\boldsymbol{x}^j)\log P_{\hat{Y}|X_S}(y|\boldsymbol{x}^j),
}
\end{equation}
where $P_{\hat{Y}|X_S}$ denotes a probabilistic hypothesis and $P_{Y_S|X_S}$ is the target distribution with the one-hot format, \emph{i.e.},

\begin{equation}
\label{target_dist}
{\textstyle
P_{Y_S|X_S}(y|\boldsymbol{x}^j) = 1 \text{\; if \;} y=y^j \text{\; otherwise \;} 0.
}
\end{equation}
The population risk and the empirical risk are defined as 
\begin{equation} 
{\textstyle
L_{\mu}(w) \triangleq \underset{(x,y)\sim\mu}{\mathbb{E}}\ell(w,\boldsymbol{x},y); \; L_{S}(w) \triangleq \frac{1}{J}\sum_{j=1}^J\ell(w, \boldsymbol{x}^j,y^j).
}
\end{equation}

In the information theoretic framework \cite{russo2015much}, $P_{W|S}$ denotes a learning algorithm  and the expected generalization error is 
$\text{gen}(\mu, P_{W|S}) \triangleq \mathbb{E}[L_{\mu}(W) - L_{S}(W)]$,
where $W$ is the random variable of $w$, and the expectation is taken with respect to $P_{S,W} = \mu^{\otimes J}\otimes P_{W|S}$ and $\otimes$ denotes the Hadamard product.

\textbf{Theorem 1} \cite{xu2017information}. If $\ell$ is $\sigma$-subgaussian, $\text{gen}(\mu, P_{W|S})$ is bounded by $I(S;W)$, \emph{i.e.},
\begin{equation}
\label{mi_boundp}
{\textstyle
|\text{gen}(\mu, P_{W|S})| \leq \sqrt{\frac{2\sigma^2}{J}I(S; W)}.
}
\end{equation}
Theorem 1 formulates the intuition that a learning algorithm would generalize better if it has less dependence on $S$.


$\text{DNNs} = \{\boldsymbol{x}; \boldsymbol{t}_1; \cdots; \boldsymbol{t}_I; \hat{\boldsymbol{y}}\}$ denote neural networks with $I$ hidden layers and is trained on $S$ by minimizing $\ell_{\text{CE}}$.
To simplify theoretical derivation, we focus on feedforward fully connected DNNs, \emph{i.e.}, MultiLayer Perceptrons (MLPs), on the supervised image classification task.
Without loss of generality, most theoretical conclusions about MLPs are based on the $\text{MLP} = \{\boldsymbol{x}; \boldsymbol{t}_1; \boldsymbol{t}_2; \hat{\boldsymbol{y}}\}$.
We denote ${T_i}$ as the random variable of $\boldsymbol{t}_i$, and $H(\cdot)$ as entropy.

In the MLP, $\boldsymbol{t}_1 =\{t_{1n} = \sigma_1[\langle \boldsymbol{\omega}^{(1)}_{n}, \boldsymbol{x} \rangle]\}_{n=1}^{N}$ consists of $N$ neurons, where $\langle \boldsymbol{\omega}^{(1)}_{n}, \boldsymbol{x} \rangle = \sum_{m=1}^M\omega^{(1)}_{mn} \cdot x_m + b_{1n}$ is the $n$th dot-product given the weight $\omega^{(1)}_{mn}$ and the bias $b_{1n}$, and $\sigma_1(\cdot)$ is an activation function.
Similarly, $\boldsymbol{t_{2}} = \{t_{2k}=\sigma_2[\langle \boldsymbol{\omega}^{(2)}_{k}, \boldsymbol{t}_1 \rangle]\}_{k=1}^K$ consists of $K$ neurons, where $\langle \boldsymbol{\omega}^{(2)}_{k}, \boldsymbol{t}_1 \rangle = \sum_{n=1}^N\omega^{(2)}_{nk} \cdot t_{1n} + b_{2k}$.
The output layer $\hat{\boldsymbol{y}} = \{\hat{y}_l = \frac{1}{Z_Y}\text{exp}[\langle \boldsymbol{\omega}^{(3)}_{l}, \boldsymbol{t}_2 \rangle]\}_{l=1}^L$ is softmax with $L$ nodes,
where $\langle \boldsymbol{\omega}^{(3)}_{l}, \boldsymbol{t}_2 \rangle = \sum_{k=1}^K\omega^{(3)}_{kl} \cdot t_{2k} + b_{yl}$ and ${Z_Y} = \sum_{l=1}^L\text{exp}[\langle \boldsymbol{\omega}^{(3)}_{l}, \boldsymbol{t}_2 \rangle]$ is the partition function.
All activation functions are $\text{ReLU}(z) = \max(0,z)$.

\section{Estimating the mutual information based on a probabilistic representation for DNNs}

In this section, we specify the mutual information estimator for $I(X_S;T)$ and $I(Y_S;T)$.
Notably, we do not adopt the classical non-parametric models, \emph{e.g.} the empirical distribution \cite{shwartz2017opening} and the kernel density estimation \cite{IP-argue} to estimate $I(X_S;T)$ and $I(Y_S;T)$, as classical non-parametric models derives poor mutual information estimation \cite{paninski2003estimation, mcallester2020formal}.
Here, we define the probability space of a hidden layer, which enables us to accurately estimate the mutual information.

\subsection{The probability space of a hidden layer}

It is known that a convolution kernel (namely the weights of convolution) defines a local feature, and a convolution operation derives a feature map to measure the cross-correlation between the local feature and input in a receptive field (see Chapter 9.1 in \cite{goodfellow2016deep}).
Notably, a fully connected layer is equivalent to a convolution layer with the kernel size having the same dimension as input. Thus the weights of a neuron can be viewed as a global feature, and a fully connected layer derives activations to measure the cross-correlation between the global features and input.


Assuming that a fully connected layer $\boldsymbol{t}$ consists of $N$ neurons $\{t_n = \sigma[\langle \boldsymbol{\omega}_n, \boldsymbol{z} \rangle]\}_{n=1}^N$, where $\boldsymbol{z} \in \mathbb{R}^M$ is the input of $\boldsymbol{t}$, $\langle \boldsymbol{\omega}_n, \boldsymbol{z} \rangle = \sum_{m=1}^M\omega_{mn}\cdot z_m + b_n$ is the dot-product between $\boldsymbol{z}$ and $\boldsymbol{\omega}_n$, and $\sigma(\cdot)$ is an activation function.
Based on the cross-correlation explanation, the behavior of $\boldsymbol{t}$ is to measure the cross-correlations between $\boldsymbol{z}$ and the $N$ possible features defined by the weights $\{\boldsymbol{\omega}_n\}_{n=1}^N$. 
In the context of pattern recognition \cite{vapnik1999overview}, we define a virtual random process or `experiment' as $\boldsymbol{t}$ recognizing one of the patterns/features with the largest cross-correlation to $\boldsymbol{z}$ from the $N$ possible features.
The experiment characterizes the behavior of $\boldsymbol{t}$ (\emph{i.e.}, before recognizing the features with the largest cross-correlation, $\boldsymbol{t}$ must measure the cross-correlations between $\boldsymbol{z}$ and all the $N$ possible features) while meets the requirement of the `experiment' definition (\emph{i.e.}, only one outcome occurs on each trial of the experiment \cite{durrett2019probability}). 
Finally, the probability space $(\Omega_T,\mathcal{F},P_T)$ is defined as follows:

\paragraph{Definition 1.} $(\Omega_T,\mathcal{F},P_T)$ consists of three components: the sample space $\Omega_T$ has $N$ possible outcomes (features) $\{\boldsymbol{\omega}_n = \{{\omega}_{mn}\}_{m=1}^{M}\}_{n=1}^N$ defined by the weights\footnote{We do not take into account the scalar value $b_n$ for defining $\Omega_T$, as it not affects the feature defined by $\boldsymbol{\omega}_n$.} of the $N$ neurons; the event space $\mathcal{F}$ is the $\sigma$-algebra; and the probability measure $P_T$ is a Gibbs distribution \cite{energy_learning} to quantify the probability of $\boldsymbol{\omega}_n$ being recognized as the feature with the largest cross-correlation to $\boldsymbol{z}$.

\begin{figure}[!t]
\centering
\vspace{-0.1in}
\includegraphics[scale=0.4]{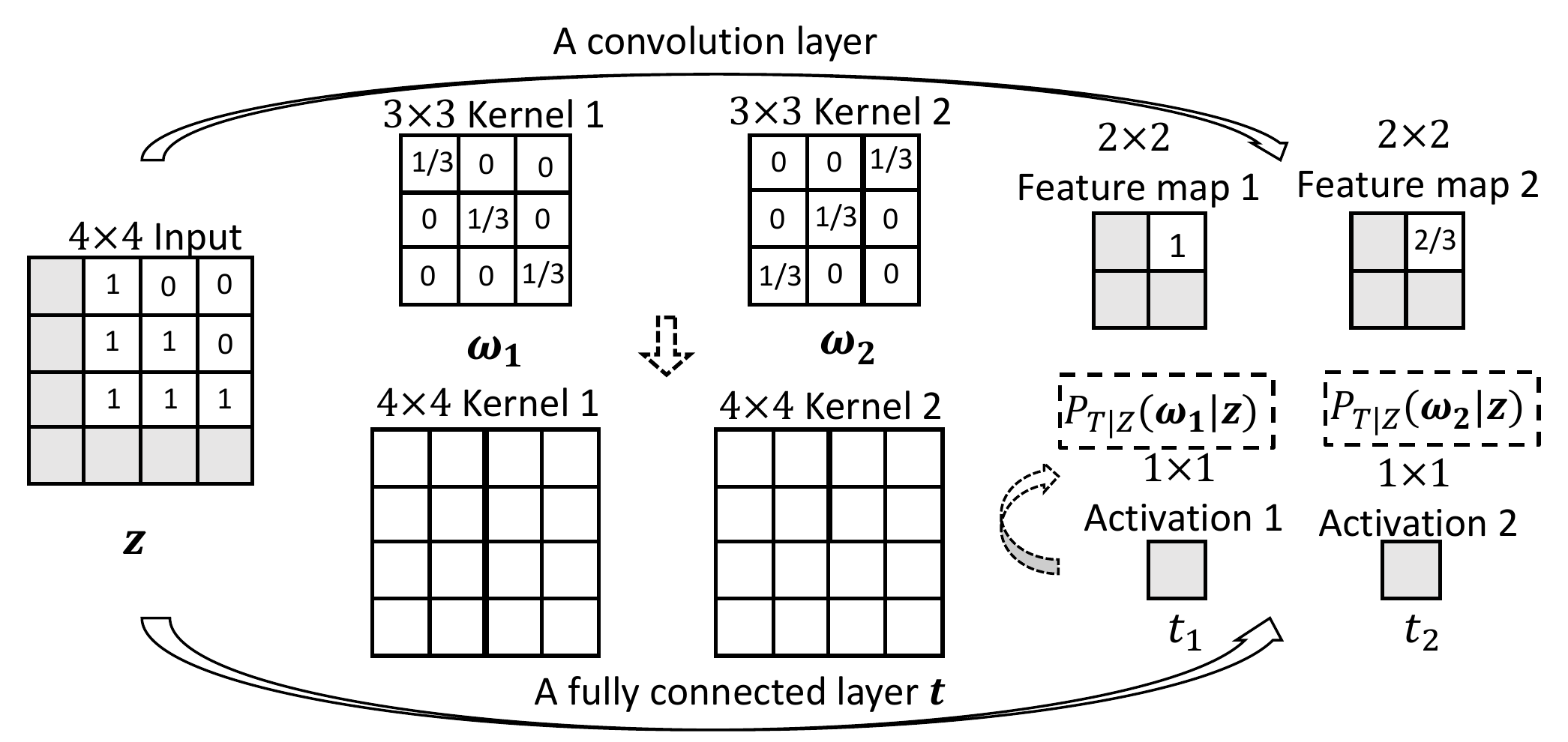}
\vspace{-0.2in}
\caption{
Given a $4\times 4$ input $\boldsymbol{z}$, a fully connected layer $\boldsymbol{t}$ is equivalent to a convolution layer with $4\times 4$ convolution kernels.
The definition of convolution implies that the $4\times 4$ weights $\boldsymbol{\omega}_1$ and $\boldsymbol{\omega}_2$ define two global features, and the two activations $t_1, t_2$ indicate the cross-correlation between $\boldsymbol{\omega}_1,\boldsymbol{\omega}_2$ and $\boldsymbol{z}$, respectively.
}
\label{Img_prob_space}
\vspace{-0.2in}
\end{figure}

Taking into account the randomness of $\boldsymbol{z}$, the conditional distribution $P_{T|Z}$ is formulated as
\begin{equation} 
\label{Gibbs_f}
\begin{split}
P_{T|Z}(\boldsymbol{\omega}_n|\boldsymbol{z}) = \frac{1}{Z_{T}}\text{exp}(t_{n}) = \frac{1}{Z_{T}}\text{exp}[\sigma(\langle \boldsymbol{\omega}_n, \boldsymbol{z} \rangle)],\\
\end{split}
\end{equation}
where $Z_{T} = \sum_{n=1}^N\text{exp}(t_{n})$ is the partition function and $Z$ is the random variable of $\boldsymbol{z}$.

$(\Omega_T, \mathcal{F}, P_T)$ clearly explains all the ingredients of $\boldsymbol{t}$ in a probabilistic fashion.
The $n$th neuron defines a global feature by the weights $\boldsymbol{w}_n$ and the activation $t_n = \sigma(\langle \boldsymbol{\omega}_{n}, \boldsymbol{z} \rangle)$ measures the cross-correlation between $\boldsymbol{w}_n$ and $\boldsymbol{z}$. 
The Gibbs distribution $P_{T|Z}$ indicates that if $\boldsymbol{w}_n$ has the larger activation, \emph{i.e.}, the larger cross-correlation to $\boldsymbol{z}$, it has the larger probability being recognized as the feature with largest cross-correlation to $\boldsymbol{z}$.
For instance, if $\boldsymbol{z} \in \mathbb{R}^{16}$ and $\boldsymbol{t}$ has $N = 2$ neurons, then $\Omega_T = \{\boldsymbol{\omega}_1, \boldsymbol{\omega}_2\}$ defines two possible outcomes (features), where $\boldsymbol{\omega}_n = \{\omega_{mn}\}_{m=1}^{16}$.
${\textstyle \mathcal{F} = \{\emptyset, \{\boldsymbol{\omega}_1\}, \{\boldsymbol{\omega}_2\}, \{\boldsymbol{\omega}_1, \boldsymbol{\omega}_2\}\}}$ means that neither, one, or both of the features are recognized by $\boldsymbol{t}$ given $\boldsymbol{z}$, respectively.
$P_{T|Z}(\boldsymbol{\omega}_1|\boldsymbol{z})$ and $P_{T|Z}(\boldsymbol{\omega}_2|\boldsymbol{z})$ are the probability of $\boldsymbol{\omega}_1$ and $\boldsymbol{\omega}_2$  being recognized as the feature with the largest cross-correlation to $\boldsymbol{z}$, respectively. 
Figure \ref{Img_prob_space} shows the probabilistic explanation for a fully connected layer.

\textbf{Definition 2.} The random variable $T: \Omega_T \rightarrow E_T$ of $\boldsymbol{t}$ is
\begin{equation}
{\textstyle
T(\boldsymbol{\omega}_n) \triangleq n.
}
\end{equation}
where the measurable space $E_T = \{1,\cdots, N\}$. 
Notably, the one-to-one correspondence between $\boldsymbol{\omega}_n$ and $n$ indicates
\begin{equation}
{\textstyle
    P_{T|Z}(\boldsymbol{\omega}_n|\boldsymbol{z}) = P_{T|Z}(n|\boldsymbol{z}).
}
\end{equation}
Based on the probability space $(\Omega_T,\mathcal{F},P_T)$, the next section specifies the MI estimator for $I(X_S;T)$ and $I(Y_S;T)$.

\subsection{The mutual information estimator}
\label{mi_estimator}

To estimate $I(X_S; T)= H(T) - H(T|X_S)$, we need to specify $P(T|X_S)$ and $P(T)$.
Based on $(\Omega_{T}, \mathcal{F}, P_{T})$, $P_{T|X_S}(n|\boldsymbol{x}^j)$ can be expressed as
\begin{equation}
\label{pfall_given_x}
\begin{split}
{\textstyle P_{T|X_S}(n|\boldsymbol{x}^j) = \frac {1}{Z_{T}}\text{exp}[\sigma(\langle \boldsymbol{\omega}_n, \boldsymbol{x}^j \rangle)].}
\end{split}
\end{equation}
To derive the marginal distribution $P(T)$,  we sum the joint distribution $P(T, X_S)$ over $\boldsymbol{x}^j\in S$,
\begin{equation}
\label{average_gibbs_fi}
\begin{split}
{\textstyle
P_T(n) = \sum_{\boldsymbol{x}^j \in S}P_{X_S}(\boldsymbol{x}^j)P_{T|X_S}(n|\boldsymbol{x}^j),
}
\end{split}
\end{equation}
where $P_{X_S}(\boldsymbol{x}^j)$ is estimated by the empirical distribution $1/J$ given $S$.
As a result, we can derive $I(X;T_i)$ by Equation (\ref{pfall_given_x}, \ref{average_gibbs_fi}).
Similarly, to estimate $I(Y_S; {T}) = H(T) - H(T|Y_S)$, we reformulate $P(T|Y_S)$ as
\begin{equation}
\label{pdf_f_given_y}
{\textstyle
P_{T|Y_S}(n|y) = \sum_{\boldsymbol{x}^j \in S}P_{T|X_S}(n|\boldsymbol{x}^j)P_{X_S|Y_S}(\boldsymbol{x}^j|y)
}
\end{equation}
where $P_{X_S|Y_S}(\boldsymbol{x}^j|y)$ is estimated by the empirical distribution $1/\mathbb{N}(y)$, and $\mathbb{N}(y)$ is the number of samples with the label $y$ in $S$.
$I(Y_S;T)$ can be derived by Equation (\ref{average_gibbs_fi},\ref{pdf_f_given_y}).

\section{Estimating the mutual information bound}


The target distribution, Equation (\ref{target_dist}), indicates the conditional entropy $H(Y_S|X_S)$ being zero, \emph{i.e.},
\begin{equation}
{\scriptstyle
    H(Y_S|X_S) = -\sum_{\boldsymbol{x}^j\in S}\sum_{y=1}^{L}P_{X_S,Y_S}(\boldsymbol{x}^j,y)\text{log}P_{Y_S|X_S}(y|\boldsymbol{x}^j)=0.
}
\end{equation}
Since $H(Y_S) = H(Y_S|X_S) + I(X_S;Y_S)$, we have 
\begin{equation}
{\textstyle
    H(Y_S) = I(X_S;Y_S).
}
\end{equation}
Since $H(X_S,Y_S) = H(X_S|Y_S) + I(X_S;Y_S) + H(Y_S|X_S)$, $H(S) = H(X_S,Y_S)$ can be expressed as
\begin{equation}
{\textstyle
    H(S) = H(\bar{X}_S) + H(Y_S) = H(X_S),
}
\end{equation}
where $\bar{X}_S$ is a virtual random variable containing all the information of $X_S$ except $Y_S$, \emph{i.e.}, $H(\bar{X}_S) = H(X_S|Y_S)$. 

\begin{figure}[t]
\vskip -0.1in
\begin{center}
\centerline{\includegraphics[scale=0.35]{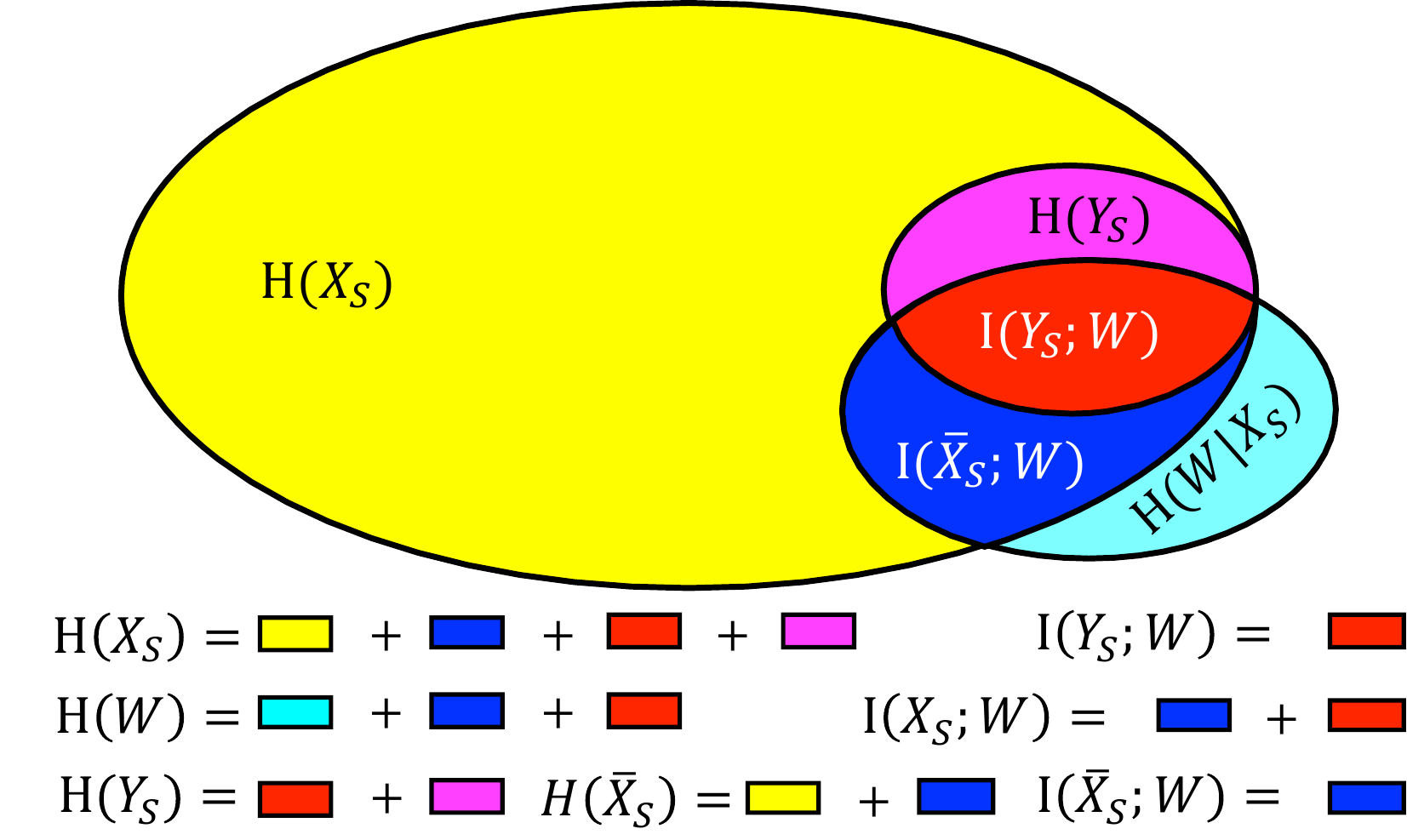}}
\vskip -0.1in
\caption{The Venn diagram of $I(X_S;W)$, $I(\bar{X}_S;W)$, $I(Y_S;W)$.}
\label{Img_venn_samples}
\end{center}
\vskip -0.4in
\end{figure}

A learning algorithm, \emph{e.g.}, the MLP, learning information from $S$ can be expressed as,
\begin{equation}
\label{mi_sw}
    I(S;W) = I(\bar{X}_S;W) + I(Y_S;W).
\end{equation}
As a result, the information theoretic relation between $S$ and $W$ can be visualized by Figure \ref{Img_venn_samples}.

A learning iteration, \emph{i.e.}, an epoch, consists of two phases: (i) training and (ii) inference.
During inference, all the weights are fixed, thus the $\text{MLP} =\{\boldsymbol{x};\boldsymbol{t}_1;\boldsymbol{t}_2, \hat{\boldsymbol{y}}\}$ forms a Markov chain $X_S \rightarrow T_1 \rightarrow T_2 \rightarrow \hat{Y}$ \cite{shwartz2017opening}.
Since $\bar{X}_S$ is a part of $X_S$, the information flow of $\bar{X}_S$ also satisfies the Markov chain
\begin{equation}
\label{Markov_X_bar}
{\textstyle
    \bar{X}_S \rightarrow T_1 \rightarrow T_2 \rightarrow \hat{Y}.
}
\end{equation}
During training, the information of $Y_S$ flows into the MLP in the backward direction.
Since $Y_S$ not has any information of $\bar{X}_S$, it will not affect the information of $\bar{X}_S$ in each layer during training.
Hence, Equation (\ref{Markov_X_bar}) characterizes the information flow of $\bar{X}$ in the MLP during both training and inference, and $I(\bar{X}_S;W)$ can be simplified as $I(\bar{X}_S;T_1)$.

It is known that a well trained DNN derives correct predictions in the output layer, \emph{i.e.,} the output layer contains all the information of labels.
As a result, $I(Y_S;W)$ can be estimated by $I(Y_S;\hat{Y})$. Overall, we can derive
\begin{equation}
\label{mi_equation}
    I(S;W) = I(\bar{X}_S;T_1) + I(Y_S;\hat{Y}).
\end{equation}
Equation (\ref{mi_equation}) explain the connection between the architecture of the MLP and $I(S;W)$, thus we can clarify the effect of MLP architecture on the generalization behavior of MLPs based on Theorem 1 (see Section \ref{exp3}).

In summary, we study the information flow in the MLP and simplify $I(S;W)$ as Equation (\ref{mi_equation}), in which $I(\bar{X}_S;T_1)$ can be derived as $I(X_S;T_1)-I(Y_S;T_1)$ by Equation (\ref{mi_sw}).
As a result, we can directly estimate $I(S;W)$ via Section \ref{mi_estimator} and avoid the limitations of relaxations.

\section{Experiments}

\subsection{Setup}

\begin{figure}[t]
\centering
\includegraphics[scale=0.4]{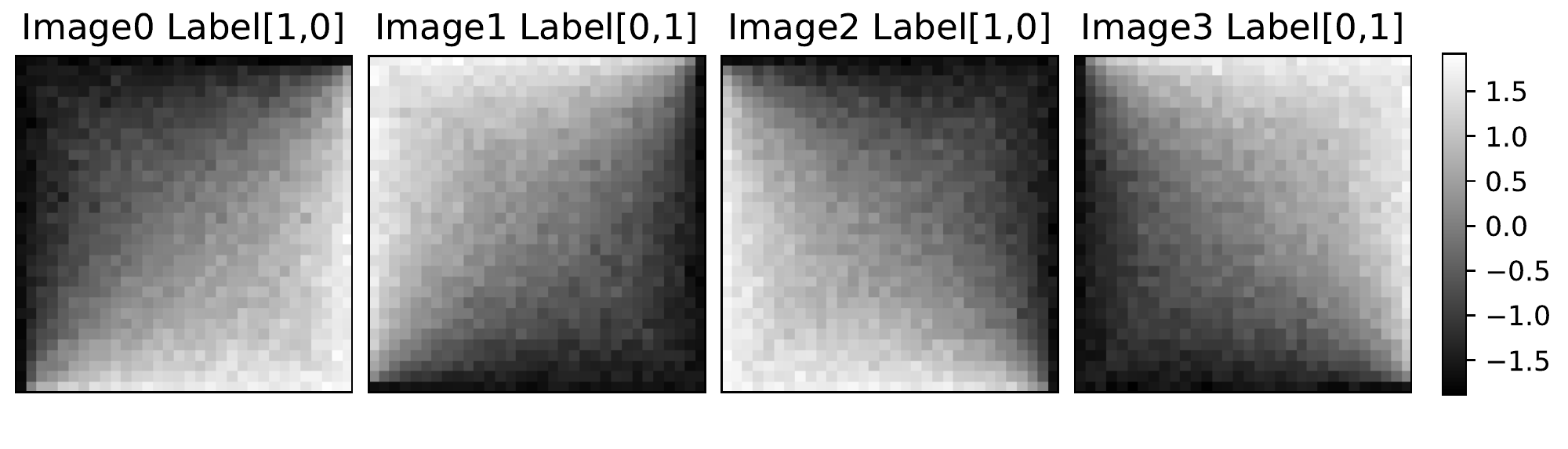}
\vspace{-0.2in}
\caption{\small{
All the images are added $\mathcal{N}(\mu, \sigma^2)$. 
Image0 is not rotated,
Image1 is generated by rotating $\hat{\boldsymbol{x}}$ along the secondary diagonal direction,
Image2 and Image3 are generated by rotating $\hat{\boldsymbol{x}}$ along the vertical and horizontal directions, respectively.
}}
\label{fig_iid_images1}
\vspace{-0.25in}
\end{figure}

\textbf{Dataset.}
The experiments are implemented on two datasets: a synthetic dataset and the Fashion-MNIST (abbr. FMNIST) dataset \cite{xiao2017fashion}.
The synthetic dataset consists of 512 gray-scale $32\times 32$ images, which are evenly generated by  rotating a deterministic image $\hat{\boldsymbol{x}}$ in four different orientations and adding Gaussian noise with expectation $\mu = \mathbb{E}(\hat{\boldsymbol{x}})$ and variance $\sigma^2 = 1$, namely $\boldsymbol{x} = r(\hat{\boldsymbol{x}}) + \mathcal{N}(\mu,\sigma^2)$,
where $r(\cdot)$ denotes the rotation method shown in Figure \ref{fig_iid_images1}. The reason for adding noise is to avoid networks to simply memorize $\hat{\boldsymbol{x}}$.
The binary labels [1,0] and [0,1] evenly divide the synthetic dataset into two classes.
As a result, the synthetic dataset has (approximately) 2 bits information and the labels have 1 bit information.
Compared to benchmark dataset, the entropy of the synthetic dataset is known, which allows us to clearly demonstrate the MI estimator.

\textbf{Reference.} We choose the latest relaxation method, namely the incoherence of gradients, as reference, and compare the generalization bound derived by the MI estimatior to the relaxed MI bound \cite{negrea2019information} and the relaxed Conditional MI (CMI) bound \cite{haghifam2020sharpened}.

\textbf{Neural Networks.} We train six MLPs on the two dataset by a variant of Stochastic Gradient Descent (SGD) method, namely Adam \cite{kingma2014adam}, over 1000 epochs with the learning rate $\alpha = 0.03$.
Table \ref{MLP3} summarizes the architecture of the six MLPs.
Following the same experimental procedures, we train the MLPs with 100 different random initialization to derive the average MI bound.
The code is available online\footnote{https://github.com/EthanLan/A-Tighter-Genearlization-Bound-in-deep-learning}.

\begin{table}[h]
\vskip -0.1in
\caption{The number of neurons in each layer of the six MLPs}
\label{MLP3}
\begin{center}
\begin{small}
\scalebox{0.8}{
\begin{tabular}{ccccc|ccccc}
\toprule
Synthetic & $\boldsymbol{x}$ & $\boldsymbol{t}_1$ & $\boldsymbol{t}_2$ & $\hat{\boldsymbol{y}}$& FMNIST  & $\boldsymbol{x}$ & $\boldsymbol{t}_1$ & $\boldsymbol{t}_2$ & $\hat{\boldsymbol{y}}$ \\
\midrule
MLP1 & 1024 & 11 & 6 & 2 & MLP4 & 784 & 512 & 256 & 10\\
MLP2 & 1024 & 7 & 6 & 2 & MLP5 & 784 & 256 & 128 & 10\\
MLP3 & 1024 & 3 & 6 & 2 & MLP6 & 784 & 32 & 16 & 10\\
\bottomrule
\end{tabular}
}
\end{small}
\end{center}
\vskip -0.2in
\end{table}

\subsection{Validating the MI estimator and the MI bound}
\label{exp3}

\begin{figure}[t]
\centering
\includegraphics[scale=0.5]{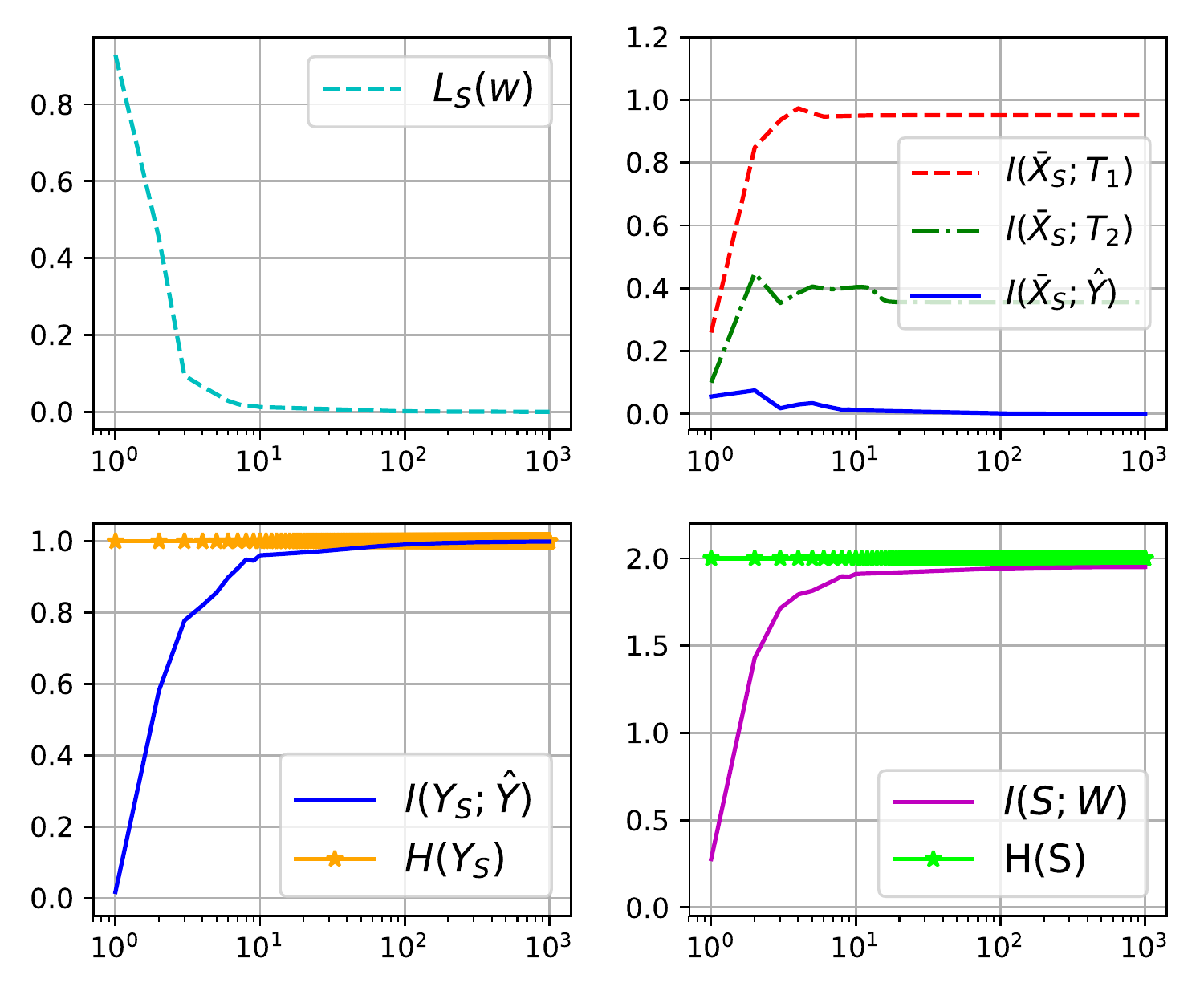}
\vspace{-0.2in}
\caption{\small{
The x-axis is the epoch index.
The figures show the variations of $L_S(w)$, $I(\bar{X}_S;T_i)$, $I(Y_S;\hat{Y})$, and $I(S;W)$ in MLP1. 
}}
\label{Img_mi_synthetic}
\vspace{-0.1in}
\end{figure}

\begin{figure*}
\centering
\vskip -0.1in
\begin{minipage}[b]{0.9\linewidth}
\centering
\centerline{\includegraphics[scale=0.5]{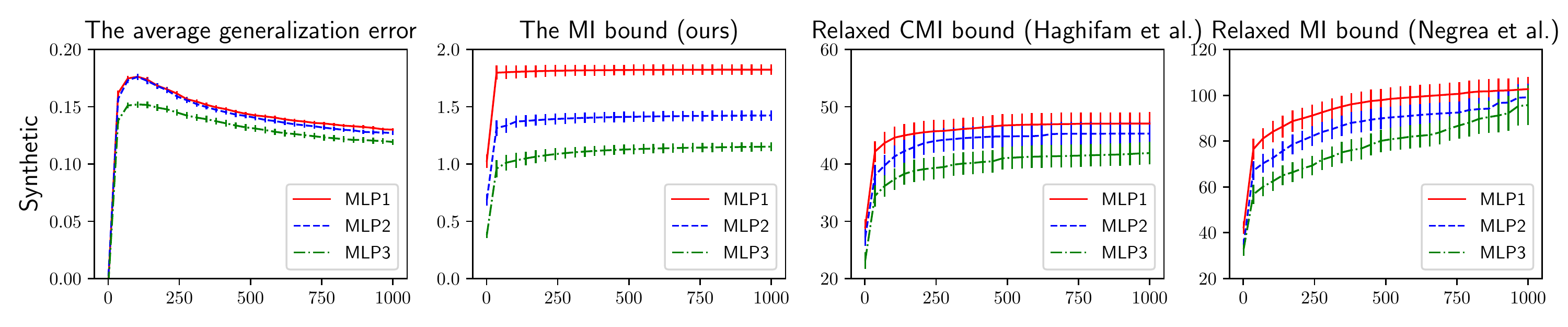}}
\end{minipage}
\centering
\begin{minipage}[b]{0.9\linewidth}
\centering
\centerline{\includegraphics[scale=0.5]{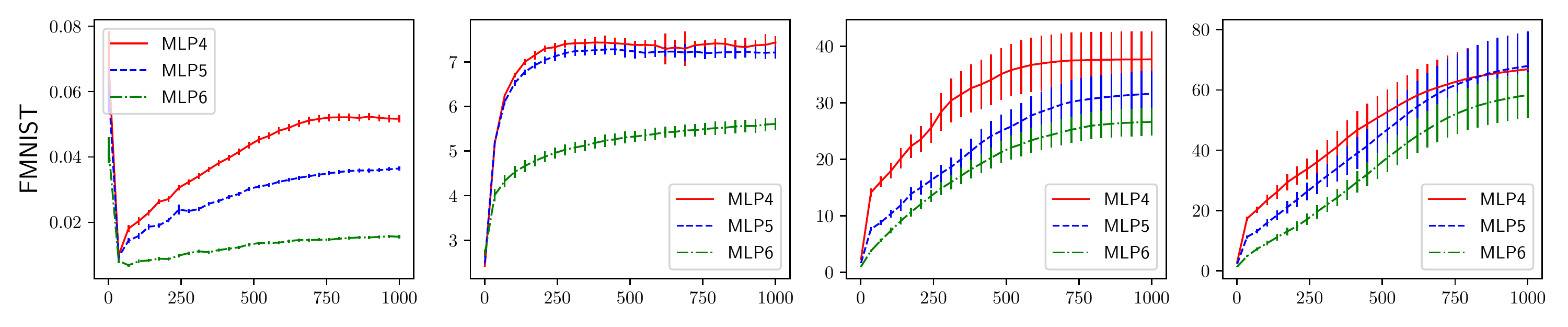}}
\end{minipage}
\vskip -0.2in
\caption{
The x-axis is the epoch index.
The first and second row show the average generalization error (i.e., training accuracy - testing accuracy), the MI bound derived by the MI estimator and the relaxation on the synthetic and FMNIST dataset, respectively.
}
\label{Img_mlp_bounds}
\vskip -0.2in
\end{figure*}

First, we demonstrate the MI estimator. 
Figure \ref{Img_mi_synthetic} show $I(\bar{X}_S;T_1) \geq I(\bar{X}_S;T_2) \geq I(\bar{X}_S;\hat{Y})$, which confirms that Equation (\ref{Markov_X_bar}) characterizes the information flow of $\bar{X}_S$.
As $L_S(w)$ decreases to zero, we observe that $I(Y_S;\hat{Y})$ converges to $H(Y_S)$, \emph{i.e.}, $I(Y_S;W)$ can be estimated by $I(Y_S;\hat{Y})$.
Finally, $I(S;W)$ converging to $H(S)$ in MLP1 validates the MI estimator, namely Equation (\ref{mi_equation}).

Second, we validate the MI explanation for generalization.
Figure \ref{Img_mlp_bounds} shows that if a MLP has more neurons (\emph{i.e.}, more complicated), it learns more information (\emph{i.e.},$I(S;W)$ is larger), which shows the same trend as the generalization error of the MLP.
Therefore, $I(S;W)$ correctly predicts the generalization behavior of the MLP.

Third, we show that the MI estimator derives a much tighter bound than the relaxation. For instance,  the MI bound derived by the MI estimator is less than 2 in MLP1, whereas the relaxed CMI bound is over 40 and the relaxed MI bound is over 80, \emph{i.e.}, the MI estimator predicts the generalization behavior more accurately than the relaxation. 
We derive similar results for other MLPs on the FMNIST dataset.

\section{Conclusion and future work}

In this work, we introduce a probabilistic representation for accurately estimating $I(S;W) = I(\bar{X}_S; T_1)+ I(Y_S; \hat{Y})$.
Based on the MI estimator, we validate the MI explanation for generalization, and show that directly estimating $I(S;W)$ derives a much tighter generalization bound than the existing relaxations.
A potential direction is to extend the work to general networks, \emph{e.g.}, CNNs.

\bibliography{it-dnn}
\bibliographystyle{icml2021}



\end{document}